\documentclass{esannV2}
\usepackage{graphicx}
\usepackage[latin1]{inputenc}
\usepackage{amssymb,amsmath,array}

\usepackage{subcaption}
\usepackage{booktabs}

\usepackage{tikz}
\usetikzlibrary{arrows.meta,positioning}

\usepackage{hyperref}

\begin{document}

\title{Out-of-Distribution Segmentation via Wasserstein-Based Evidential Uncertainty}

\author{Arnold Brosch$^1$, Abdelrahman Eldesokey$^2$, Michael Felsberg$^3$ and Kira Maag$^1$
%
%
\vspace{.3cm}\\
%
1- Heinrich-Heine-University D\"usseldorf, Germany, \{arnold.brosch,kira.maag\}@hhu.de
%
\vspace{.1cm}\\
2- KAUST, Saudi Arabia, abdelrahman.eldesokey@kaust.edu.sa 
\vspace{.1cm}\\
3- Link\"oping University, Sweden, michael.felsberg@liu.se
 \\
}

\maketitle

\begin{abstract}
Deep neural networks achieve superior performance in semantic segmentation, but are limited to a predefined set of classes, which leads to failures when they encounter unknown objects in open-world scenarios. Recognizing and segmenting these out-of-distribution (OOD) objects is crucial for safety-critical applications such as automated driving. In this work, we present an
evidence segmentation framework using a Wasserstein loss, which captures distributional distances while respecting the probability simplex geometry. Combined with Kullback-Leibler regularization  
and Dice structural consistency terms,
our approach leads to improved OOD segmentation performance compared to uncertainty-based approaches. 
\end{abstract}

%
%
%
\section{Introduction}
Deep neural networks (DNNs) have achieved remarkable success in computer vision tasks such as semantic segmentation, which is the task of assigning a class label to each pixel of an image from a fixed and predefined set of semantic classes \cite{Chen2018_deeplab}. 
In safety-critical domains such as automated driving, semantic segmentation enables robust scene understanding.
However, 
DNNs show a significant decrease in performance when used in open-world environments where unseen objects occur that are not present in the training distribution. 
This challenge of out-of-distribution (OOD) segmentation~\cite{Chan2021_smiyc} aims to identify and segment objects that do not belong to known classes to avoid dangerous situations. 
Many OOD segmentation methods are uncertainty-based techniques that do not require auxiliary models or OOD data. 
DNNs quantify uncertainty using the softmax output, employing common uncertainty metrics such as the maximum softmax probability or entropy \cite{Hendrycks2016}, whereby the predicted class probability is interpreted as a measure of confidence. However, these softmax-based measures are often 
tend to produce overly confident predictions, especially for OOD samples. 
Bayesian approximations such as Monte Carlo (MC) Dropout \cite{Mukhoti2020} or deep ensembles \cite{Lakshminarayanan2017} attempt to overcome this limitation by estimating prediction uncertainty through model sampling, 
but they require considerable computational effort during inference due to multiple stochastic forward passes.

In this work, we propose an evidential deep learning (EDL) framework using Wasserstein loss for uncertainty-aware semantic segmentation. 
EDL encourages the model to express higher uncertainty in unfamiliar or ambiguous regions of the input space. 
Typically, EDL uses Euclidean objectives like expected mean squared error (MSE)-based loss as it  
works directly on the expected values of the evidence distribution, but leads to overconfident predictions \cite{Sensoy2018_edl}. 
Therefore, we propose using a Wasserstein loss as it captures the distance between distributions more accurately in metric terms, i.e., the Wasserstein distance accounts for the underlying geometry of the probability simplex, thereby enabling more smooth and calibrated uncertainty estimates. 
This novel loss formulation combines a distribution-aware Wasserstein term with Kullback-Leibler regularization, which 
prevents overconfidence for OOD inputs. 
In addition, Dice 
is included to 
enforce spatial coherence in the segmentation output. 
Together, these components result in a model that improves OOD segmentation compared to uncertainty-based baselines. 
The code is publicly available on \url{https://github.com/A-Brosch/EDL-OOD-Segmentation}.

%
%
%
\section{Related Work}
Existing approaches for OOD segmentation can be grouped into four main research directions, (i) methods relying on auxiliary models to detect anomalies (e.g., reconstruction networks \cite{Vojir2023}), (ii) approaches leveraging additional OOD or synthetic data during training \cite{Delic2024}, 
(iii) techniques operating in the feature-space to identify outliers \cite{Sodano2024}, and (iv) uncertainty-based methods that quantify the model's confidence. 
The latter includes approaches such as maximum softmax probability \cite{Hendrycks2016} and foreground-background segmentation 
confidence \cite{Marschall2025}, as well as sampling-based methods such as MC Dropout \cite{Mukhoti2020} and deep ensembles \cite{Lakshminarayanan2017}. 
More advanced uncertainty-based approaches go beyond output scores and require access to the model's gradients \cite{Maag2024_grads} or intermediate computations \cite{Liang2018}.

Most works rely on auxiliary models, require additional OOD data, or operate in feature-space, which increases complexity and limits general applicability. 
In contrast, we present an uncertainty-based method in which the confidence of the model is estimated directly from its predictions, without the need for additional data or auxiliary architectures. 
Compared to other uncertainty-based techniques, our approach does not rely on sampling or access 
backward passes, making it most comparable to simpler approaches such as maximum softmax. 

%
%
%

%
\section{Method}
\emph{Evidential Deep Learning.} \hspace{0.3ex} 
In semantic segmentation, each pixel $z$ of an input image $x$ is assigned a class label $y$ from the set of classes $i \in \{1, \ldots, C\}$.
The DNN predicts per-pixel (pre-activation) logits, which are transformed into uncertainty-aware outputs using EDL~\cite{Sensoy2018_edl}, which has also been applied to semantic segmentation in \cite{Holmquist23}.
Instead of applying a softmax function to obtain class scores (single-point predictions), the network produces 
\emph{evidence values} $e_i$ for each class using ReLU. 
These evidences 
define the parameters of a Dirichlet distribution over class probabilities, $\alpha_i = e_i + 1$, $S = \sum_{i=1}^{C} \alpha_i $,
which represents a higher-order distribution capturing both prediction and uncertainty. 
The estimated probability for class $i$ and the corresponding 
\emph{uncertainty} are given by $p_i = \frac{\alpha_i}{S}$ and  $\mathcal{U} = \frac{C}{S}$. 
Here, $S$ represents the total collected evidence, reflecting the model's overall confidence in its pixel-wise prediction. 
The idea is that there is 
high uncertainty for classes that not included in the training data, enabling identification of OOD objects.

\begin{figure*}[t]
\centering
\begin{minipage}[t]{0.6\textwidth}
    \vspace{0pt} 
    \centering
    \captionof{table}{OOD segmentation results for RoadAnomaly21.}
    \label{tab:ood_anomaly}
    \resizebox{\linewidth}{!}{
    \begin{tabular}{cccc ccccc}
    \toprule 
    \multicolumn{4}{c}{} & \multicolumn{5}{c}{RoadAnomaly21} \\
    \cmidrule(r){5-9}
    & & & &
    AuPRC $\uparrow$ &
    FPR$_{95}$ $\downarrow$ &
    $\overline{\text{sIoU}}$ $\uparrow$ &
    $\overline{\text{PPV}}$ $\uparrow$ &
    $\overline{F_1}$ $\uparrow$
    \\
    \midrule
    \multicolumn{4}{l}{Ensemble}         & 17.7 & 91.1 & 16.4 & \textbf{20.8} & 3.4 \\
    \multicolumn{4}{l}{MC Dropout}       & 28.9 & \textbf{69.5} & \textbf{20.5} & 17.3 & 4.3 \\
    \multicolumn{4}{l}{Maximum Softmax}  & 28.0 & 72.1 & 15.5 & 15.3 &\textbf{5.4} \\
    \multicolumn{4}{l}{Entropy}          & \textbf{30.0} & 73.0 & 17.8 & 15.6 & 5.1 \\
    \midrule
    $\lambda_{\mathcal{W}}$ & $\lambda_{\text{Dice}}$ & $\lambda_{\text{KL}}$ & $\lambda_{\text{MSE}}$ &
    \multicolumn{5}{c}{} \\
    \cmidrule(r){1-4}
    0.00 & 0.00 & 0.00 & 1.00  & 33.4 & 68.8 & 19.0 & 17.4 & 3.8 \\
    1.00 & 0.00 & 0.00 & 0.00  & 39.7 & 61.8 & 14.5 & 21.6 & 4.0 \\   
    \midrule
    1.00 & 0.00 & 0.00 & 0.40  & \textbf{54.3} & 50.4 & 26.3 & \textbf{21.5} & 7.8 \\
    1.00 & 0.00 & 0.00 & 0.45  & 50.5 & 49.6 & \textbf{28.0} & 20.8 & \textbf{9.0} \\
    1.00 & 0.00 & 0.00 & 0.50  & 40.0 & 53.1 & 19.1 & 20.3 & 4.6 \\
    \midrule
    1.00 & 0.70 & 0.00 & 0.45  & 42.8 & \textbf{43.6} & 23.9 & 19.2 & 6.3 \\
    1.00 & 0.75 & 0.00 & 0.45  & 41.4 & 56.3 & 26.3 & 16.0 & 7.1 \\
    1.00 & 0.80 & 0.00 & 0.45  & 37.9 & 65.5 & 22.9 & 15.7 & 4.8 \\
    \midrule
    1.00 & 0.75 & 0.10 & 0.45  & 35.0 & 48.2 & 16.3 &  6.6 & 5.0 \\
    1.00 & 0.75 & 0.15 & 0.45  & 43.6 & 53.6 & 24.6 & 15.2 & 5.9 \\
    1.00 & 0.75 & 0.20 & 0.45  & 35.4 & 46.9 & 20.6 & 17.5 & 5.6 \\
    \bottomrule
    \end{tabular}
    }
\end{minipage}
\hfill
\begin{minipage}[t]{0.39\textwidth}
    \vspace{0pt} 
    \centering

     \resizebox{0.65\linewidth}{!}{\tikzset{
  corner/.style    = {circle, fill=black, inner sep=0.9pt},
  sampleMSE/.style = {circle, fill=blue!65,   inner sep=0.8pt},
  sampleW/.style   = {circle, fill=orange!80, inner sep=0.8pt},
  pullMSE/.style   = {->, line width=0.45pt, opacity=0.7, shorten >=1.5pt, shorten <=6.5pt, draw=blue!65},
  pullW/.style     = {->, line width=0.45pt, opacity=0.7, shorten >=1.5pt, draw=orange!80},
  tri/.style       = {line width=0.4pt},
  axislabel/.style = {font=\scriptsize}
}

\newcommand{\DrawSmallTriangle}{
  \coordinate (A) at (0,0);
  \coordinate (B) at (3,0);
  \coordinate (C) at (1.5,2.598); 
  \draw[tri] (A)--(B)--(C)--cycle;
  \node[corner,label=below left:{\scriptsize class 1}] at (A) {};
  \node[corner,label=below right:{\scriptsize class 2}] at (B) {};
  \node[corner,label=above:{\scriptsize class 3}] at (C) {};
}

\begin{tikzpicture}[scale=0.8]
  \DrawSmallTriangle

  \foreach \x/\y in {0.5/0.2,0.7/0.25,2.3/0.25,2.5/0.2,1.45/2.2,1.55/2.2}
    {\node[sampleMSE] at (\x,\y) {}; }

  \draw[pullMSE] (1.5,1.0) -- (0.7,0.25);
  \draw[pullMSE] (1.6,1.1) -- (2.3,0.25);
  \draw[pullMSE] (1.5,1.2) -- (1.5,2.2);

  \foreach \x/\y in {1.3/1.0,1.7/1.0,1.4/1.2,1.6/1.3,1.5/1.1,1.2/1.3,1.8/1.4}
    {\node[sampleW] at (\x,\y) {}; }

  \draw[pullW] (1.0,1.1) -- (1.4,1.1);
  \draw[pullW] (2.0,1.1) -- (1.6,1.2);
  \draw[pullW] (1.5,0.8) -- (1.5,1.1);

\end{tikzpicture}}
    \vspace{0.3em}
    
    \vspace{0.5em}

    \includegraphics[width=0.9\linewidth]{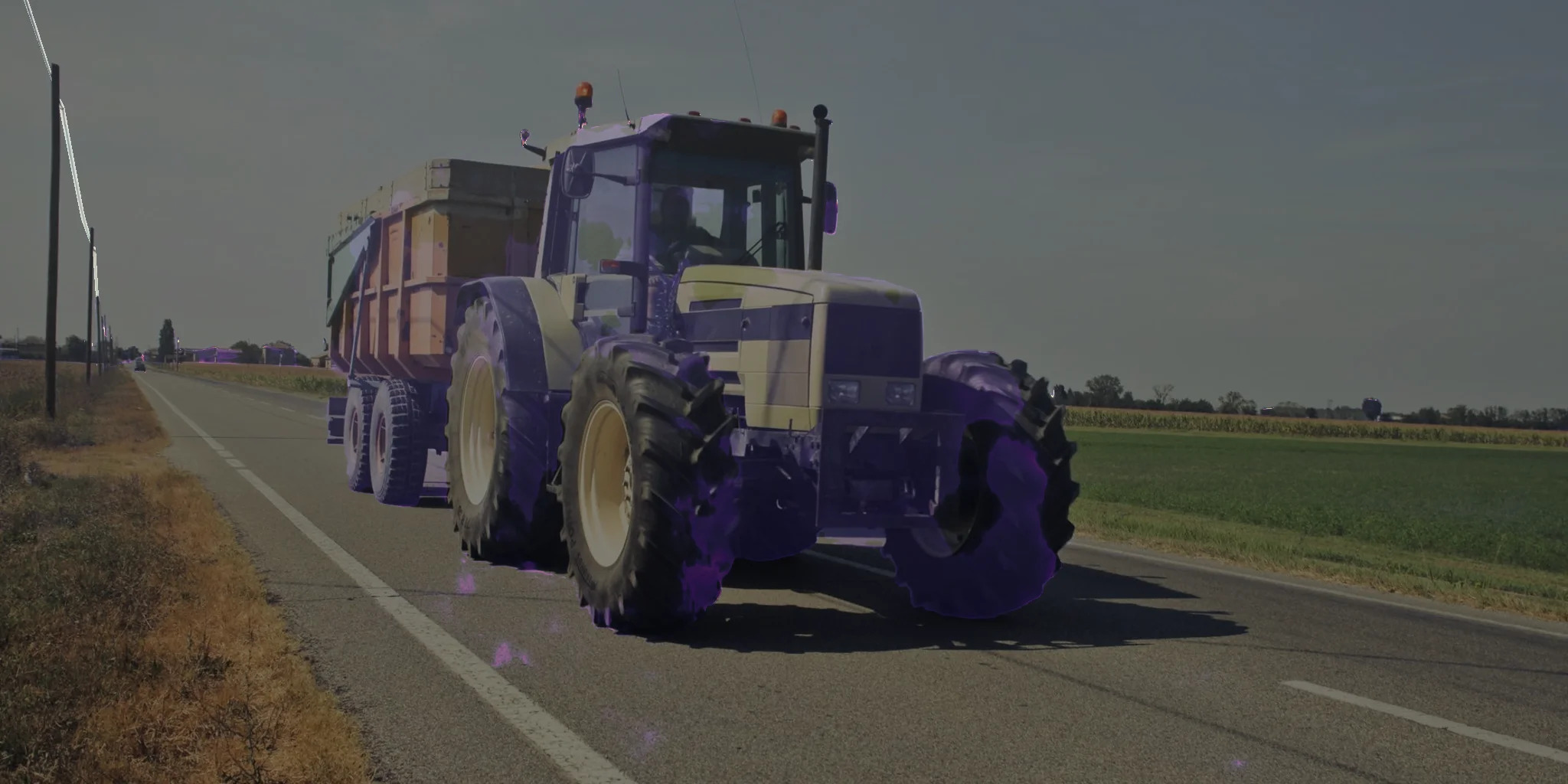}
    \vspace{0.3em}

    \vspace{-1.0em}

    \caption{
        Comparison of MSE (\emph{blue}) and Wasserstein loss (\emph{orange}) on the three-class probability simplex (\emph{top})
        and an uncertainty heatmap of the AnomalyTrack (\emph{bottom}).
    }
    \label{fig:simplex_examples}
\end{minipage}
\end{figure*}

\emph{Geometric Intuition + Loss Construction.} \hspace{0.3ex} 
Most existing EDL formulations employ Euclidean objectives like expected MSE. 
The MSE loss, which incorporates the Dirichlet variance is represented by $\mathcal{L}_{\text{MSE}_i} 
= (p_i - y_i)^2 + \frac{p_i(1 - p_i)}{S + 1}$,
where $y_i$ denotes the value for the $i$-th class of one-hot encoded ground truth vector.
This formulation enables the reduction of prediction error simultaneous to reducing the variance. 
The MSE loss is beneficial for obtaining sharp segmentation for in-distribution by pushing predictions near the simplex vertices (class boundaries), corresponding to overconfident one-hot outputs, see Figure~\ref{fig:simplex_examples} (top). 

However, this behavior contradicts the detection of OOD pixels, as we are interested in maintaining a state of uncertainty for OOD objects. 
To capture the distance between predictive and target distributions in a geometrically meaningful way, we employ a Wasserstein loss formulation. 
Instead of pushing predictions towards the vertices as MSE, the Wasserstein loss does not shift uncertainties to classes but centers them between the boundaries (see Figure~\ref{fig:simplex_examples} (top)). 
Using the Dirac metric as the transport cost, the Wasserstein distance between the predicted class probability distribution and the ground truth one-hot distribution reduces to the probability assigned to the true class, 
$\mathcal{L}_{\mathcal{W}} 
= \mathbb{E}\!\left[\,1 - p_y\,\right] $, 
where $p_{y}$ represents the predicted mean probability for the correct class $y$.

While the Wasserstein loss enforces distribution alignment and improves uncertainty calibration, it does not explicitly consider the spatial structure of the segmentation results. To ensure that neighboring pixels belonging to the same object have consistent predictions,
we add a Dice loss term, 
$\mathcal{L}_{\text{Dice}_i} = 1 - \frac{2 \sum_{z} p_i(z) y_i(z)}{\sum_{z} p_i(z) + \sum_{z} y_i(z)}$ working on image-level, 
which measures the overlap between predicted and true segmentation masks. 

To further prevent the model from becoming overconfident in regions of high uncertainty or OOD inputs, we employ an annealed Kullback-Leibler (KL) regularization term $ \mathcal{L}_{\text{KL}} 
 = \mathrm{KL}\!\left[\mathrm{Dir}(\alpha)\,||\,\mathrm{Dir}(1)\right] $.
The KL term is gradually introduced during training through an annealing schedule $\lambda_{\text{KL}}$, allowing the network to focus on learning discriminative features before uncertainty calibration is regulated.

Our loss function is composed of
$\mathcal{L}_{\text{total}} = 
\lambda_{\mathcal{W}} \frac{1}{Z} \sum_z \mathcal{L}_{\mathcal{W}} + 
\lambda_{\text{Dice}} \frac{1}{C} \sum_{i=1}^C \mathcal{L}_{\text{Dice}_i} +
\lambda_{\text{KL}} \frac{1}{Z} \sum_z \mathcal{L}_{\text{KL}} +
\lambda_{\text{MSE}} \frac{1}{Z} \sum_z \sum_{i=1}^C \mathcal{L}_{\text{MSE}_i} $
where $Z$ is the number of pixels, and $\lambda_{\mathcal{W}}$, $\lambda_{\text{Dice}}$, $\lambda_{\text{KL}}$ and $\lambda_{\text{MSE}}$ are the different weightings. 
Although the Wasserstein loss serves as the primary objective, we retain a component of the MSE with a small weighting factor to stabilize the early training phase.
In practice, the Wasserstein objective may lead to slower convergence or unstable gradients, especially when the predicted evidence distributions are not yet calibrated. 
In summary, Wasserstein loss promotes distributional accuracy by aligning predictive and target distributions in a geometrically consistent manner, the Dice loss enforces structural accuracy through spatial and boundary coherence, 
annealed KL regularization ensures uncertainty calibration by penalizing overconfident evidence in ambiguous regions, and the MSE term contributes training stability by providing a reliable gradient signal during 
optimization.

%
%

%
%
\section{Experiments}

\begin{table*}[t]
\caption{OOD segmentation results for LostAndFound and RoadObstacle21. }
\centering
\resizebox{0.98\linewidth}{!}{
\begin{tabular}{cccc cc ccc cc ccc}
\toprule 
\multicolumn{4}{c}{} & \multicolumn{5}{c}{LostAndFound test-NoKnown} & \multicolumn{5}{c}{RoadObstacle21} \\ 
\cmidrule(r){5-9} \cmidrule(r){10-14}
& & & & AuPRC $\uparrow$ & FPR$_{95}$ $\downarrow$ & $\overline{\text{sIoU}}$ $\uparrow$ & $\overline{\text{PPV}}$ $\uparrow$ & $\overline{F_1}$ $\uparrow$
& AuPRC $\uparrow$ & FPR$_{95}$ $\downarrow$ & $\overline{\text{sIoU}}$ $\uparrow$ & $\overline{\text{PPV}}$ $\uparrow$ & $\overline{F_1}$ $\uparrow$ \\
\midrule
\multicolumn{4}{l}{Ensemble}         & 2.9 & 82.0 & 6.7 & 7.6 & 2.7      & 1.1 & 77.2 & 8.6 & 4.7 & 1.3 \\
\multicolumn{4}{l}{MC Dropout}       & 36.8 & 35.6 & 17.4 & 34.7 & 13.0 & 4.9  & 50.3 & 5.5  & 5.8  & 1.1 \\
\multicolumn{4}{l}{Maximum Softmax}  & 30.1 & 33.2 & 14.2 & \textbf{62.2} & 10.3 & 15.7 & \textbf{16.6} & \textbf{19.7} & 15.9 & 6.3 \\
\multicolumn{4}{l}{Entropy}          & \textbf{47.1} & \textbf{21.6} & \textbf{30.7} & 42.1 & \textbf{30.2} & \textbf{28.4} & 26.7 & 14.7 & \textbf{20.5} & \textbf{9.7} \\
\midrule
$\lambda_{\mathcal{W}}$ & $\lambda_{\text{Dice}}$ & $\lambda_{\text{KL}}$ & $\lambda_{\text{MSE}}$ & \multicolumn{10}{c}{} \\
\cmidrule(r){1-4}
0.00 & 0.00 & 0.00 & 1.00   & 25.7 & 56.8 & 25.9 & 27.7 & 14.6 & 1.6  & 62.1 & 18.5 & 5.6  & 2.0 \\ 
1.00 & 0.00 & 0.00 & 0.00   & 47.5 & 27.4 & 21.5 & 39.2 & 19.7 & 4.1  & 66.0 & 6.9  & 8.5  & 2.0 \\ 
\midrule
1.00 & 0.00 & 0.00 & 0.40   & 50.4 & 25.8 & 21.2 & 38.0 & 19.2 & \textbf{30.7} & 42.3 & 11.6 & 26.4 & \textbf{8.7} \\ 
1.00 & 0.00 & 0.00 & 0.45   & 53.2 & 23.2 & 25.3 & 37.7 & 20.7 & 20.2 & \textbf{31.2} & 11.0 & 18.3 & 5.8 \\ 
1.00 & 0.00 & 0.00 & 0.50   & 44.4 & 30.5 & 25.2 & \textbf{42.8} & 23.6 & 12.2 & 32.6 & 3.6  & \textbf{34.0} & 3.4 \\ 
\midrule
1.00 & 0.70 & 0.00 & 0.45 & 39.3 & 36.1 & 23.4 & 41.0 & 20.9 & 11.7 & 58.7 & \textbf{17.2} & 14.9 & 7.3\\ 
1.00 & 0.75 & 0.00 & 0.45 & 47.4 & \textbf{17.6} & \textbf{26.6} & 41.1 & \textbf{24.2} & 8.8 & 35.3 & 12.1 & 14.2 & 5.3\\ 
1.00 & 0.80 & 0.00 & 0.45 & 27.0 & 39.8 & 14.7 & 25.3 & 8.7 & 3.3 & 63.1 & 5.0 & 9.1 & 1.3\\ 
\midrule
1.00 & 0.75 & 0.10 & 0.45 & 42.1 & 25.5 & 19.8 & 36.2 & 16.3 & 2.8 & 51.1 & 10.4 & 7.4 & 2.7 \\ 
1.00 & 0.75 & 0.15 & 0.45 & \textbf{53.6} & 31.6 & 23.4 & 38.8 & 22.1 & 2.8 & 51.1 & 10.4 & 7.4 & 2.7 \\ 
1.00 & 0.75 & 0.20 & 0.45 & 49.1 & 38.2 & 21.9 & 37.5 & 20.1 & 4.9 & 39.0 & 16.2 & 5.4 & 2.7 \\ 
\bottomrule
\end{tabular} }
\label{tab:ood_obstacle}
\end{table*}
%

%
%
\subsection{Experimental Setting}
\emph{Datasets.} \hspace{0.3ex} 
For training, we use the Cityscapes dataset \cite{Cordts2016} for semantic segmentation of urban street scenes consisting of $2,\!975$ training 
images. 
For evaluation, we consider the \emph{SegmentMeIfYouCan} benchmark \cite{Chan2021_smiyc} using the 
LostAndFound (LAF, \cite{Pinggera2016}) dataset ($1,\!203$ images with small road obstacles),
RoadObstacle21 ($412$ images similar to LAF with greater diversity in OOD objects and situations), and RoadAnomaly21 ($100$ images with various unique anomalies). 

\emph{Model Training.} \hspace{0.3ex} 
We employ the DeepLabV3+ architecture \cite{Chen2018_deeplab} with ResNet-50 backbone, which is mostly used in OOD segmentation approaches. 
Training is conducted with a batch size of $4$ for $80$K iterations,
a learning rate of $3 \cdot 10^{-4}$, a weight decay of $10^{-4}$ and ADAM optimizer. 
We run ablations on the weighting parameters $\lambda$ of our combined loss function in the following.  
The KL weight is increased linearly from $0$ to $\lambda_{\text{KL}}$ between iterations $40$k and $48$k and then kept stable,  
to encourage learning for higher learning rate
and regularization when the learning rate has already dropped. We achieved an average mIoU of $72.17\%$.

\emph{Evaluation Metrics.} \hspace{0.3ex} 
To evaluate OOD segmentation performance, we follow the official benchmark protocol, i.e., use area under the precision-recall curve (AuPRC) and false positive rate at $95\%$ true positive rate (FPR$_{95}$) on pixel-level, and averaged adjusted mIoU (sIoU), positive predictive value (PPV) and $F_1$-score on segment-level.

%
%
\subsection{Numerical Results}

The results for LAF and RoadObstacle21 are given in Table~\ref{tab:ood_obstacle}, and for RoadAnomaly21 in Table~\ref{tab:ood_anomaly}. 
Firstly, we compare the use of clean MSE ($\lambda_{\mathcal{W}}=0$, $\lambda_{\text{MSE}}=1$) loss vs.\ Wasserstein ($\lambda_{\mathcal{W}}=1$, $\lambda_{\text{MSE}}=0$) and observe that, across all datasets, Wasserstein loss achieves significantly better results than MSE. 
Secondly, we added MSE loss with a small weighting to the Wasserstein loss to increase segmentation performance without losing latent uncertainty. This combination delivers significantly higher performance than if only one factor is considered. 
From now on, we will set $\lambda_{\text{MSE}}=0.45$ as it has achieved the best results for LAF and this dataset consists of the most images and therefore serves as a reference value. 
Thirdly, we are investigating the impact of DICE loss, which has a particularly positive effect on the FPR$_{95}$ metric. We will fix $\lambda_{\text{Dice}}$ to $0.75$. 
Lastly, we include annealed KL regularization to reduce overconfidence. 
We have found that a value of $\lambda_{\text{KL}}=0.15$ works best for LAF (corresponding to a small expected calibration error of $0.0347$ on in-distribution data). 
Moreover, we experimented with using lower evidence for the prior distribution (e.g., $\mathrm{Dir}(0.25)$), which weakened its dampening effect on the estimate used in the KL regularization, but this did not lead to measurable improvements.
Overall, our experiments show that we achieve the best performance with the combined loss when at least three components have an influence on the prediction.

The baseline methods against which we compare our approach are listed at the top of both tables. We use comparable methods, i.e., uncertainty-based approaches, distinguishing between sampling-based (Ensemble, MC Dropout) and output-based (Maximum Softmax, Entropy) techniques.  
We obtain improvement results compared to these baselines, especially with RoadAnomaly21, where we outperform all others.

%
%
%
\section{Conclusion}
In this work, we have introduced a Wasserstein-based evidential framework for OOD-aware semantic segmentation that jointly optimizes distribution accuracy, structural consistency, and calibrated uncertainty.  
Our method better accounts for the geometry of the probability simplex and provides smoother, more reliable uncertainty estimates than MSE. 
In the SegmentMeIfYouCan benchmark, our approach demonstrates improved performance compared to other uncertainty-based baseline methods, proving that principled uncertainty modeling enables more robust and reliable predictions in open-world environments.


\begin{footnotesize}

\bibliographystyle{unsrt}
\bibliography{biblio}

\end{footnotesize}


\end{document}